\documentclass{article}
\usepackage[utf8]{inputenc}
\usepackage{url}
\usepackage{floatrow}
\usepackage{caption}
\usepackage{hyperref}
\usepackage{graphicx}
\usepackage{verbatim}
\usepackage{xspace}
\usepackage{subcaption}
\usepackage[dvipsnames]{xcolor}
\usepackage[export]{adjustbox}
\usepackage[dvipsnames]{xcolor}
\usepackage{amsmath,amssymb}
\usepackage[authoryear]{natbib}

\usepackage{graphicx,calc}
\usepackage{float} 
\newlength\myheight
\newlength\mydepth
\settototalheight\myheight{Xygp}
\usepackage{adjustbox}

\newcommand{\projectName}{\large Ego-Exo4D Human Meshes Dataset:\\ 4D Human Motion Reconstruction for Ego-Exo Captures\xspace}

\title{\projectName{}}
\author{
  \small Abhiram Maddukuri$^{1}$ \quad  \small Georgios Pavlakos$^{1}$ \\
  \small $^{1}$The University of Texas at Austin
}

\begin{document}

\renewcommand{\arraystretch}{1.3}

\date{}
\maketitle


\begin{abstract}
\label{sec:abstract}
Ego-Exo4D is a large-scale dataset providing synchronized egocentric and multi-view exocentric video, a rich resource for skill learning and assessment, procedural activity understanding, and embodied AI. However, the dataset ships with only sparse 3D human pose annotations, and reconstructing dense human motion from its multi-view captures is nontrivial. To this end, we present Ego-Exo4D-HM, a large-scale dataset of 4D human motion reconstructions for Ego-Exo4D's captures, and release the accompanying reconstruction pipeline. The code, dataset, and documentation can be found at \url{https://abhiram824.github.io/egoexo4d_human_meshes/}.
\end{abstract}

\begin{figure}[ht]
  \centering
  \includegraphics[width=\linewidth]{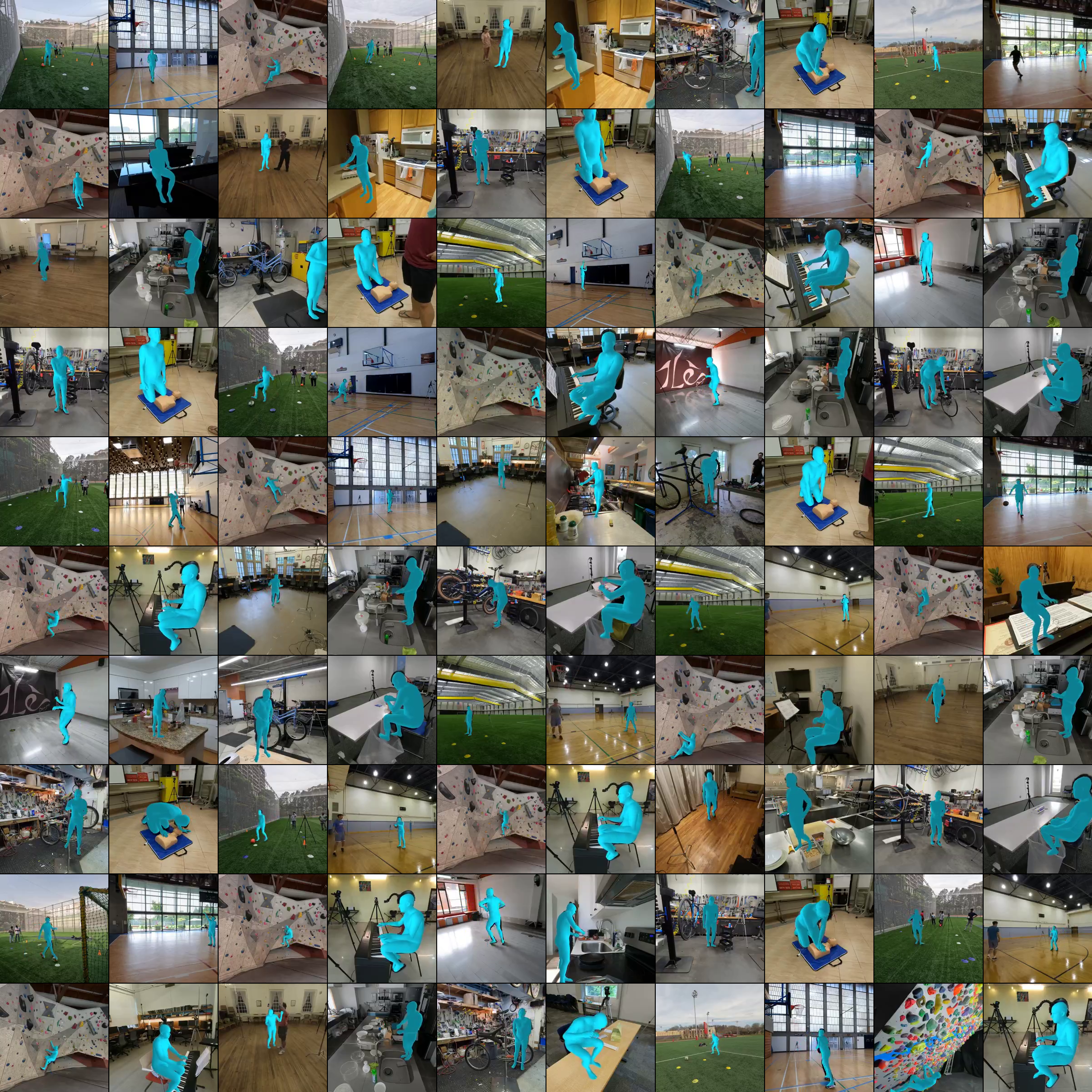}
  \caption{Recovered SMPL-H meshes overlaid on exocentric video across the reconstructed Ego-Exo4D sequences.}
  \label{fig:teaser}
\end{figure}
\section{Introduction}
\label{sec:intro}

Ego4D~\cite{grauman2022ego4d} and Ego-Exo4D~\cite{grauman2024egoexo4dunderstandingskilledhuman} are among the largest efforts in egocentric data collection, providing thousands of hours of video across a wide range of activities and participants. Datasets of this scale open the door to pipelines for skill learning and assessment, procedural activity understanding, embodied AI and robot imitation learning, and coaching or tutoring systems that give feedback on physical performance.

Ego-Exo4D in particular is a rich dataset, with extensive capture including an egocentric sensor alongside multiple exocentric cameras, all synchronized and calibrated. This multi-view setup should, in principle, significantly improve the quality of 3D human perception, since the exocentric cameras observe the full body of the person performing the activity. However, the dataset's existing 3D annotations are sparse, and current 3D human reconstruction systems can be brittle at this scale.

To enable broader use of the dataset, we develop a pipeline for recovering the 4D human motion of Ego-Exo4D's captures. Our approach builds on state-of-the-art methods for 3D human pose reconstruction~\cite{ye2023decouplinghumancameramotion, goel2023humans4d, pavlakos2023reconstructinghands3dtransformers}, adapted to take advantage of Ego-Exo4D's synchronized multi-camera setup. We release the resulting SMPL-H motion sequences as the Ego-Exo4D-HM dataset (see Figure~\ref{fig:teaser}), along with our processing pipeline, so future work can use this data directly rather than reprocessing the raw captures.
\section{Background}
\label{sec:related}

Our pipeline builds on video-based human mesh recovery methods that turn per-frame estimates, such as those from HMR2.0~\cite{goel2023humans4d}, into temporally consistent 4D motion. Methods like WHAM~\cite{shin2024wham} and GVHMR~\cite{shen2024gvhmr} do this in a feed-forward fashion; however, it is challenging to incorporate information from multiple views this way. Instead, we adopt an optimization-based formulation that naturally accepts additional constraints. Specifically, we build on SLAHMR~\cite{ye2023decouplinghumancameramotion}, which jointly optimizes pose and root trajectory using 2D body keypoints and a learned motion prior.
Our pipeline is related to prior frameworks, like EasyMocap~\cite{easymocap}, which operates with multiple calibrated views, but also considers egocentric captures and relies on more recent human pose estimation approaches.
MAMMA~\cite{cuevasvelasquez2026mammamarkerlessautomatic} performs markerless multi-view motion capture directly from raw video streams. We show a short comparison in Section~\ref{sec:results}.

Previous work has independently considered the problem of egocentric body pose recovery. EgoEgo~\cite{li2023egoego} relies only on the SLAM trajectory of the head-mounted camera, without access to image observations, while more recent work, EgoAllo~\cite{yi2025egoallo}, incorporates hand observations estimated from egocentric images. The exocentric cameras of Ego-Exo4D simplify the problem in this setting, since they are observing the body from multiple views.

Our work is heavily motivated by growing use of human motion and activity data in embodied AI. Many works \cite{Luo_2026, pmlr-v305-ze25a, li2025amo, pmlr-v270-he25b} have leveraged diverse motion capture data of human activity to train performant humanoid whole-body controllers. Another line of work \cite{kareer2025egomimic, pmlr-v305-qiu25a, li2026matters, shi2025zeromimic, pmlr-v270-li25a} uses RGB videos of humans doing tasks to learn visuomotor manipulation policies. Human activity videos have also been useful in learning expressive action-conditioned video models and world models \cite{gao2026dreamdojo, goswami2026worldmodelslearningdexterous, bai2026whole}. We hope that our dataset can be useful in advancing these research directions.

Beyond embodied AI, understanding and modeling human activity is also central to skill assessment, coaching, and video forecasting. Previous work has used estimates of 3D motion in egocentric settings to provide actionable feedback~\cite{ashutosh2025expertaf}, predict future interactions~\cite{ashutosh2025fiction}, forecast hand motion~\cite{hatano2025invisible}, and edit novice motion toward an expert's skill level~\cite{somayazulu2026expertedit}. We anticipate that populating ego-exo captures with dense 3D motion estimates at scale will be valuable for supporting these methods.
\section{Methods}
  \label{sec:methods}

Given synchronized egocentric and exocentric video from the Ego-Exo4D activity dataset~\cite{grauman2024egoexo4dunderstandingskilledhuman}, our goal is to reconstruct world-frame 3D body and hand motion of the person performing the activity. We build off SLAHMR~\cite{ye2023decouplinghumancameramotion}, which can be naturally extended to multi-view settings. 
Following SLAHMR, we represent the person's state at timestep $t$ as:
\begin{equation}
    \mathcal{P}_t = \{\Phi_t, \Theta_t, \beta, \Gamma_t\}
\end{equation}
where $\Phi_t \in SO(3)$ is the global root orientation, $\Theta_t \in \mathbb{R}^{J \times 3}$ encodes the body pose across $J$ joints, $\beta \in \mathbb{R}^{16}$ is the time-invariant body shape, and $\Gamma_t \in \mathbb{R}^3$ is the root translation at timestep $t$. From here we use the SMPL-H \cite{Romero_2017} model to generate the mesh vertices $\mathbf{V}_t \in \mathbb{R}^{3\times6890}$ and joints $\mathbf{J}_t \in \mathbb{R}^{3\times67}$ of a human body through the function $\mathcal{M}$:
  \begin{equation}
      \label{eqn:obs_frame}
      [\mathbf{V}_t, \mathbf{J}_t] = \mathcal{M}(\Phi_t, \Theta_t, \beta) + \Gamma_t.
  \end{equation}

Our mesh recovery pipeline modifies SLAHMR~\cite{ye2023decouplinghumancameramotion} to leverage Ego-Exo4D's exocentric calibrations and egocentric SLAM estimates. Concretely, our pipeline consists of 3 stages: (1) Single-view pose estimation, (2) Triangulation, and (3) SMPL-H Optimization.

\subsection{Single-view pose estimation}
Takes frequently contain bystanders, so we first identify the camera wearer in each exocentric view. We run a Mask R-CNN~\cite{he2018maskrcnn} detector with a RegNetY-4GF~\cite{radosavovic2020designingnetworkdesignspaces} backbone and keep the detection whose box contains the wearer's projected 3D head position, given by the Aria SLAM trajectory. For each selected box we run ViTPose~\cite{xu2022vitposesimplevisiontransformer} to obtain human body keypoints and use HaMeR~\cite{pavlakos2023reconstructinghands3dtransformers} to obtain hand keypoints. In total we get 67 hand and body keypoints.

\subsection{Triangulation}
Each of the 67 keypoints is triangulated independently per frame by minimizing reprojection error over all calibrated views, using nonlinear least squares over the 3D point; at least two views are required. For the hand and wrist keypoints, which are more prone to occlusion and misdetection, we additionally run RANSAC~\cite{10.1145/358669.358692} over views to reject outlier detections.

\subsection{SMPL-H Optimization}
The final stage adapts SLAHMR's optimization to the calibrated multi-view setting. The free variables are the per-frame global translation $\Gamma_t$, root orientation $\Phi_t$, body and hand pose $\Theta_t$, and a single shape vector $\beta$ per take. All camera parameters are frozen to their calibrated values, so the motion is recovered directly in the metric Ego-Exo4D world frame. We utilize the same optimization process as SLAHMR~\cite{ye2023decouplinghumancameramotion}, but do not use the motion prior as the triangulated 3D evidence already constrains global motion.

\section{Results}
\label{sec:results}

\subsection{Curated Dataset}
We ran our pipeline on a subset of approximately 3,200 videos from the Ego-Exo4D dataset. From here we apply a quality filter based on two per-take criteria.
\begin{enumerate}
    \item \emph{Reprojection self-consistency}: we reproject the final optimized 3D joints through the calibrated cameras and measure the pixel distance to the corresponding 2D detections; a take is flagged if more than 10\% of these samples exceed 50\,px.

    \item \emph{Triangulation coverage}: a take is flagged if fewer than 50\% of its keypoint observations were successfully triangulated, which catches takes reconstructed from too few agreeing views. 
\end{enumerate}

  \begin{figure}[ht]
    \centering
    \includegraphics[width=\linewidth]{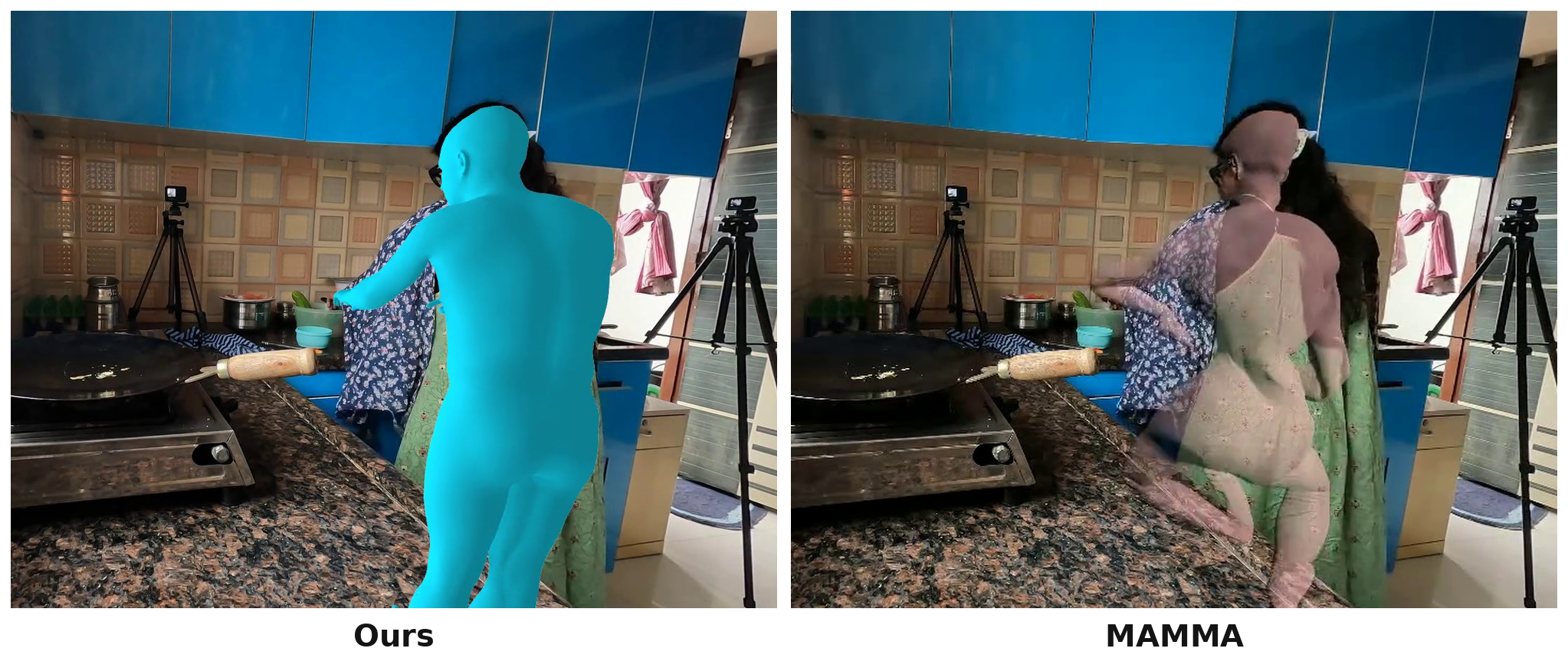}
    \caption{Comparison of our pipeline with MAMMA~\cite{cuevasvelasquez2026mammamarkerlessautomatic}.}
    \label{fig:mamma_comparison}
  \end{figure}
  
A take failing either criterion is discarded, removing 551 takes (17.1\%) and retaining 2,649. Ego-Exo4D-HM totals 104.59 hours of reconstructed motion; since each take provides four exocentric views and an egocentric view, the dataset totals 522.96 hours of video. For each take we release \texttt{npz} files containing the optimized SMPL-H parameters $(\Gamma_t, \Phi_t, \Theta_t, \beta)$, the calibrated per-frame camera intrinsics and world-to-camera extrinsics, and the 3D joints $\mathbf{J}_t$ together with their 2D reprojections in each view, from which the mesh vertices $\mathbf{V}_t$ and keypoints in any view can be regenerated via Eq.~\eqref{eqn:obs_frame}.

\subsection{Quantitative Results}
We evaluate reconstruction accuracy against the ground-truth 3D body and hand keypoint annotations provided by Ego-Exo4D. Since our reconstructions live directly in the dataset's metric world frame, we report global MPJPE, i.e.\ the mean Euclidean distance between predicted and annotated 3D joints without any alignment. Our method achieves a global MPJPE of 56.21\,mm for body keypoints (over 845 annotated takes) and 51.59\,mm for hand keypoints (over 190 annotated takes).

\subsection{Comparison with MAMMA}
We additionally compare qualitatively with MAMMA~\cite{cuevasvelasquez2026mammamarkerlessautomatic}, a recent multi-view mesh recovery method. On the partial body views, occlusions, and truncations common in Ego-Exo4D's captures, we find that our pipeline produces more robust reconstructions than MAMMA; see Fig.~\ref{fig:mamma_comparison}.

\vspace{1em}

\noindent
\textbf{Acknowledgements:} This project received computing support on the Lonestar6 GPU Cluster through the Center for Generative AI (CGAI) and the Texas Advanced Computing Center (TACC) at the University of Texas at Austin.

\bibliographystyle{plain}
\bibliography{references}

\end{document}